\documentclass[10pt,twocolumn,letterpaper]{article}
\pdfoutput=1

\usepackage{times}
\usepackage{epsfig}
\usepackage{graphicx}
\usepackage{amsmath}
\usepackage{amssymb}

\usepackage{amsmath,amsfonts,bm}

\def\Figref#1{Figure~\ref{#1}}

\def\eqref#1{equation~\ref{#1}}
\def\Eqref#1{Equation~\ref{#1}}

\def\Algref#1{Algorithm~\ref{#1}}

\def\1{\bm{1}}

\def\vzero{{\bm{0}}}

\def\vmu{{\bm{\mu}}}

\def\vsigma{{\bm{\sigma}}}

\def\vc{{\bm{c}}}

\def\vf{{\bm{f}}}

\def\vh{{\bm{h}}}

\def\vr{{\bm{r}}}

\def\vt{{\bm{t}}}
\def\vu{{\bm{u}}}
\def\vv{{\bm{v}}}

\def\vx{{\bm{x}}}
\def\vy{{\bm{y}}}
\def\vz{{\bm{z}}}

\DeclareMathAlphabet{\mathsfit}{\encodingdefault}{\sfdefault}{m}{sl}
\SetMathAlphabet{\mathsfit}{bold}{\encodingdefault}{\sfdefault}{bx}{n}

\newcommand{\Ls}{\mathcal{L}}

\newcommand{\reg}{\lambda}

\newcommand{\softmax}{\mathrm{softmax}}

\usepackage{multirow}
\usepackage{colortbl}
\usepackage{wrapfig}
\usepackage{subcaption}
\newcommand{\pointrot}{\odot}
\newcommand{\cprod}{\circ}
\newcommand{\Tabref}[1]{Table~\ref{#1}}
\usepackage{algorithm}
\usepackage{appendix}
\usepackage[noend]{algpseudocode}

\newcommand{\citet}[1]{\cite{#1}}
\newcommand{\Appref}[1]{Appendix \ref{#1}}
\newcommand{\citep}[1]{\cite{#1}}

\usepackage[pagebackref=true,breaklinks=true,letterpaper=true,colorlinks,bookmarks=false]{hyperref}

\begin{document}

\title{Geometric Capsule Autoencoders for 3D Point Clouds}
\author{Nitish Srivastava, Hanlin Goh, Ruslan Salakhutdinov\\
Apple Inc.\\
\texttt{\{nitish\_srivastava,hanlin,rsalakhutdinov\}@apple.com}
}
\date{}

\maketitle

\begin{abstract}
We propose a method to learn object representations from 3D point clouds using bundles of geometrically interpretable hidden units, which we call \mbox{``geometric capsules''}.
Each geometric capsule represents a visual entity, such as an object or a part, and consists of two components: a pose and a feature.
The pose encodes ``where'' the entity is, while the feature encodes ``what'' it is.
We use these capsules to construct a Geometric Capsule Autoencoder that learns to group 3D points into parts (small local surfaces), and these parts into the whole object, in an unsupervised manner.
Our novel Multi-View Agreement voting mechanism is used to discover an object's canonical pose and its pose-invariant feature vector. 
Using the ShapeNet and ModelNet40 datasets, we analyze the properties of the learned representations and show the benefits of having multiple votes agree.
We perform alignment and retrieval of arbitrarily rotated objects -- tasks that evaluate our model's object identification and canonical pose recovery capabilities -- and obtained insightful results.

\end{abstract}

\section{Introduction}

\begin{figure}[t]
  \centering
  \begin{center}
  \includegraphics[width=\linewidth, trim=273 188 267 242, clip]{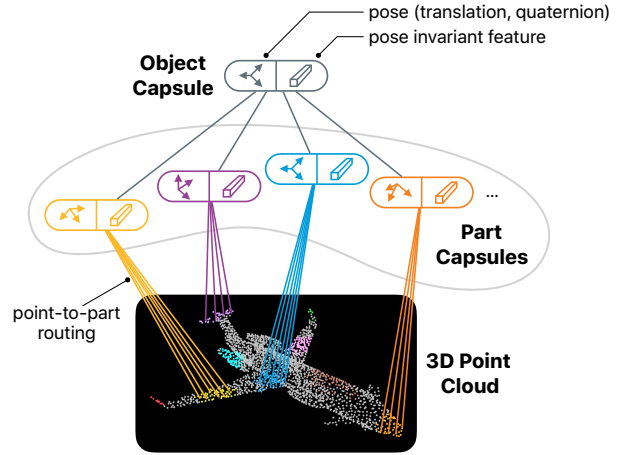} 
  \end{center}
  \vspace{-1em}
\caption{\small Model Overview. Geometric Capsules are used to represent parts, as well as the whole object. Each capsule consists of two components: pose and feature. Points are routed to parts, as shown above. The object capsule is computed from the part capsules.}
  \vspace{-1em}
\label{fig:intro}
\end{figure}

Capsule networks structure hidden units into groups, called capsules~\citep{hinton11}.
Each capsule represents a single visual entity and its hidden units collectively encode all the information regarding the entity in one place.
For example, the length of the hidden unit vector can represent the existence of the entity and its direction can represent the entity's instantiation parameters (``pose")~\citep{dynamicrouting}.
Capsule networks combine the expressive power of distributed representations (used within each capsule) with the interpretability of having one computational entity per real-world entity.

These capsules can be organized in a hierarchical manner to encode a visual scene.
Low-level capsules can be used to represent low-level visual entities (such as edges or object parts), while high-level capsules may represent entire objects.
A routing algorithm~\citep{dynamicrouting, e2018matrix} is used to discover the connections between the low-level and high-level capsules.
This makes it easy to introduce priors, such as ``one part can only belong to one object" by enforcing a mutually exclusive routing of a part capsule to a single object capsule.
These priors would be hard to enforce for visual representations that do not have such explicit grouping, such as a layer of hidden units in a convolutional neural network (CNN).

We propose a geometric capsule design, in which every visual entity (i.e. part or whole object) is encoded using two components: a pose and a feature, as shown in~\Figref{fig:intro}. 
These encode ``where'' the object is and ``what'' the object is respectively.
An entity's pose is {\em explicitly} represented in a {\em geometrically interpretable} manner, as a 6 degree-of-freedom (DOF) coordinate transformation, which encodes the canonical pose of that entity with respect to the viewer. 
The feature is represented as a real-valued vector which encodes all non-pose attributes of the object (such as its shape) and is meant to be invariant to the object's pose with respect to the viewer.
A Geometric Capsule Autoencoder is constructed to group 3D points into parts (small local surfaces), and these parts into the whole object in a completely unsupervised manner. We introduce a novel Multi-View Agreement voting mechanism within the autoencoder, which enables the concurrent discovery of an object's canonical pose and its pose-invariant feature vector.

\section{Related Work}
\label{sec:relatedwork}


Hinton et al.~\citep{hinton11} introduced the notion of capsules with
Transforming Autoencoders where the autoencoder was tasked with explicitly
computing the pose of the input data, along with a pose-invariant feature.
Interest in capsule networks was revived by Sabour et
al.~\citep{dynamicrouting} who proposed a dynamic routing algorithm for capsules.
Since then, several variants of capsule design and
routing algorithms have been proposed for various domains, such as handwritten
digits~\citep{dynamicrouting,e2018matrix,kosiorek2019stacked},
images~\citep{e2018matrix,kosiorek2019stacked} and 3D point
clouds~\citep{3dcapsnet, qenet}.

\vspace{-1em}
\paragraph{Modeling geometric pose.}
In \citep{dynamicrouting}, a capsule's existence and pose are
encoded by its hidden unit vector magnitude and direction respectively.  The
pose vector models a space of variability and is not geometrically
interpretable, especially for higher-level capsules.  \citep{e2018matrix} used $4\!\times\!4$ matrices to mediate capsule
interactions but did not constrain them to represent geometrically
interpretable transforms. \citet{kosiorek2019stacked} explicitly represents 2D affine
transforms using $3\!\times\!3$ transformation matrices.  However, instead of
extending their approach to a 3D setting, we use \emph{quaternions} to model
pose because they are easily constrained to represent only 3D rotations,
compared to $4\!\times\!4$ matrices. \cite{qenet} also uses quaternions to encode the pose of capsules.

\vspace{-1em}
\paragraph{Voting and Agreement}
An important aspect of capsule network design is how to infer the state of a
parent capsule. One approach is to have each
part vote for the state of the parent by transforming its own state via a
learned weight matrix~\citet{dynamicrouting,e2018matrix}. The votes are aggregated to form a consensus, which determines the state of the parent capsule. Parts that agree with the consensus are then assigned to that parent. This works in the
setting where discretely many parent objects are being considered for
existence. In that case, each part casts a vote per object and can form a
reasonably good vote since the vote is conditioned on the index of the object.
However, if we want to represent generic objects (i.e. objects are value-coded
instead of place-coded), it seems intuitively unreasonable to expect a part to
come up with a good vote for its parent object, since the same part can belong
to multiple objects. While each part constrains the space of possible objects, this constraint is difficult to specify as a single vote.
A different way to do this is to simply feed all the parts collectively into a Set Transformer~\citep{pmlr-v97-lee19d} to output the parent's state.
Since the Transformer has simultaneous access to all the parts, it can compute the parent's feature representation directly without explicit voting or agreement among the parts.
\citet{kosiorek2019stacked} follows this approach in their encoder but during decoder a separate decoder network per object index is used, making the feature representation place-coded.
In this work, we show how a value-coded representation can be learned using Multi-View Agreement, where instead of asking parts to agree, we ask viewpoints to agree on object pose and feature.

\paragraph{Multi-view consistency.} The idea of using multiple views to come up
with a coherent understanding of an object has been studied extensively
\cite{NIPS2016_6206, mvcTulsiani18, drcTulsiani17, lin2018learning, garg2016unsupervised,
monodepth17, zhou2017unsupervised, zhu2017rethinking, gwak2017weakly,
NIPS2016_6600}. Most relevant to our work is the work of Tulsiani et
al.~\citep{mvcTulsiani18} who proposed to learn the 3D shape and canonical pose
of objects using pairs of 2D image views to generate a self-supervised
``consistency'' learning signal. In their model, CNNs are used to output the
shape and pose of the input object, starting from image inputs.  Our model also
outputs the shape and pose of the object, but in a interpretable way, creating
a parsing of the object into parts, each with their own shape and pose.


\vspace{-1.028em}
\paragraph{Capsules for 3D point clouds.}
The 3D Point Capsule Network~\citep{3dcapsnet} showed the possibility of
extracting semantically meaningful parts from 3D point clouds using
capsule-based representations.  However, the model only learns part-level
capsules and does not model an entire object as a single capsule.  Also, pose
and feature components are not disentangled, and the feature identity is
place-coded.  Recently, Quaternion Equivariant Capsule Networks~\citep{qenet}
were proposed to learn a pose-equivariant representation of objects. The model
builds a hierarchy of Local Reference Frames, where each frame is modeled as a
quaternion. However, the model is trained using supervision from class labels and relative poses.
In our model, we adopt an entirely unsupervised approach to learn entity representations from 3D point clouds.
The model decouples the representation of pose and feature, as a translation-vector and a quaternion-vector tuple, and a value-coded real-valued vector, respectively.

\section{Geometric Capsule Autoencoder}
We now describe the Geometric Capsule Autoencoder, which builds hierarchical object representations from Geometric Capsules. We show how Multi-View Agreement can be used to learn a capsule's pose and feature components.
\subsection{Geometric Capsules}
Each geometric capsule $\vc=(\vc_q,\vc_f)$ describes a visual entity with
two components: pose $\vc_q$ and feature
$\vc_f$. The pose $\vc_q$ represents the transformation between
a global frame and the entity's canonical frame. 
Each $\vc_q=(\vt, \vr)$ consists of a translation $\vt \in
\mathbb{R}^3$ and a quaternion that represents rotation $\vr \in \mathbb{R}^4, ||\vr||=1, r_0 \geq 0$.
Therefore $\vc_q$ is a 7-D vector (but with only 6-DOF) that encodes ``where" the visual entity is with respect to a global frame.
The feature $\vc_f \in \mathbb{R}^D$ represents the
identity of the entity. It defines ``what" this entity is and encodes all
attributes of the entity other than its pose. 

Geometric capsules may be observed from different viewpoints.
A viewpoint~$\vz$ is encoded by a coordinate transformation
(using a 7-D representation similar to a capsule's pose) that maps points
expressed in the viewpoint's frame to the global frame. A point $\vx^i$ observed
from a viewpoint $\vz$, will be represented as $\vx^i | {\vz} = \vz ^{-1}
\pointrot \vx^i$, where $\pointrot$ denotes the application of a transformation
to a point. A capsule $\vc = (\vc_q, \vc_f)$ observed from a viewpoint $\vz$ will
be represented as $\vc |{\vz} = (\vz ^{-1} \cprod \vc_q, \vc_f)$ where
$\cprod$ denotes the composition of two coordinate transformations. In general,
the notation ``$|{\vz}$" should be read as ``as observed from $\vz$".

\subsection{Multi-View Agreement}
\label{sec:multiview}

\begin{figure}[t]
  \begin{center}
  \includegraphics[width=\linewidth]{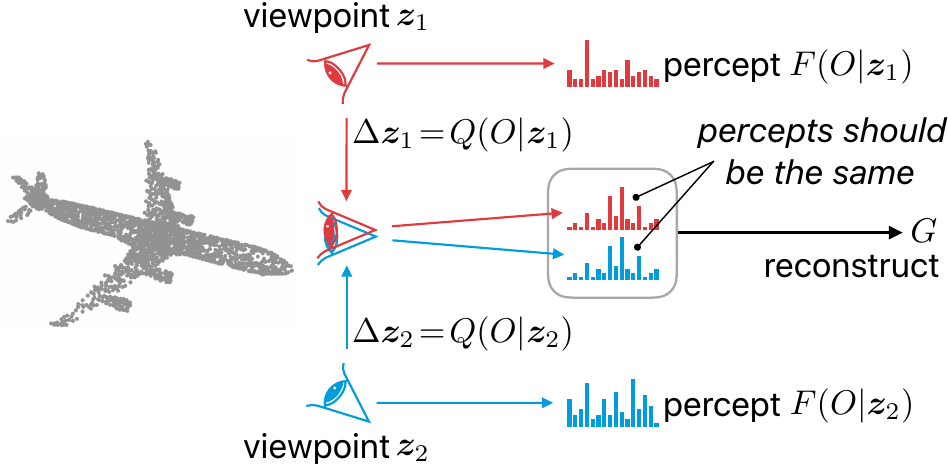}  
  \end{center}
    \vspace{-1em}
  \caption{\small Multi-View Agreement voting mechanism. A given object $O$ is viewed from random viewpoints $\vz_1, \vz_2$. The pose-voting network $Q$ computes a transform that moves each viewpoint to a canonical frame. The percepts computed by the feature-voting network $F$ from each canonical frame must now be the same.}
    \label{fig:multiview}
\end{figure}

We introduce a novel voting mechanism to discover the parent object's
value coded representation using auxiliary viewpoints. Recall that the goal is
to find the two capsule components: the parent object's canonical pose with respect to
the viewer ($\vc_q$) and a pose-invariant feature vector ($\vc_f$). Let $O$ represent an object in some global reference frame (e.g. a set of points, or a set of part capsules that belong to the object). Suppose we
have a function $F$ that records the appearance of the object (its ``percept"),
i.e. for an object $O$ and a viewpoint $\vz$ (expressed in the same global reference frame), $F(O|{\vz})$
represents the object's percept. Now suppose the object is observed from multiple
random viewpoints $Z=\{\vz_1, \vz_2, \ldots, \vz_K\}$. The percepts
$F(O|{\vz_k})$ are likely to be different, as shown in \Figref{fig:multiview}.
However, if we can find a coordinate transformation
$\Delta \vz_k=Q(O|{\vz_k})$ such that the composition $\vz_k \cprod \Delta \vz_k$, is a
canonical pose of the object, then the percepts from these transformed
viewpoints will agree (i.e. $\vf_k = F(O|{\vz_k \cprod \Delta \vz_k})$ will be the
same for all $k$).  Once this happens, the agreed
upon percept can be set to be $\vc_f$, and any one of the transformed viewpoints, say,
$\vz_1 \cprod \Delta \vz_1$ can be set to be $\vc_q$.
Note that we want the \emph{percepts} from the transformed viewpoints to agree, and not
the transformed viewpoints to themselves agree.  This mitigates issues arising
from objects having symmetries which make it impossible to have a unique
canonical pose.

In this framework, $F$ can be seen as a feature-voting network and $Q$ as a
pose-voting network, both of which will be learned.
Since the object's canonical pose is not known, no direct supervision is
available for training $Q$ (i.e. we don't have target values for $\Delta
\vz_k$).  Moreover, due to objects having symmetries, we cannot ask the
transformed viewpoints $\vz_k \cprod \Delta \vz_k$ to agree.  However, we know
that the percept votes $\vf_k$, must all agree.  Therefore, by training the
models to make the percept votes agree, both $F$ and $Q$ can be jointly
learned.  In addition to agreement, we also need to ensure that the agreed upon
percept $\vc_f$, contains all the non-pose attributes of the object.
Therefore, we train a decoder network $G$ that takes $\vc_f$ as input and
reconstructs the object in its canonical pose $O|{\vc_q} = G(\vc_f)$.

To model agreement from the percepts, we compute their mean $\vmu\!=\!\!\frac{1}{K}\!
\sum_{k=1}^K \vf_k$ and variance $\vsigma\!=\!\!\frac{1}{K}\!
\sum_{k=1}^K (\vf_k\!-\!{\vmu})^2$. The consensus percept $\vc_f$, is modeled as a
sample from $\mathcal{N}(\vmu, \vsigma^2)$. This is similar to the approach
taken in EM Routing \cite{e2018matrix}. If the votes disagree, the sample will
be noisy, and the decoder will not be able to decode the object back from the
sampled percept. Therefore, the decoder will try to reduce the noise in the
input, making the percepts agree.

\subsection{PointsToParts Autoencoder}
\label{sec:pointstoparts}
An autoencoder is used to encode a set of 3D points into a set of part-level geometric capsules.
Let $X=\{\vx^i\}_{i=1}^I$ be a set of 3D points, where $\vx^i\in\mathbb{R}^3$, and
$V=\{(\vv_q^j, \vv_f^j)\}_{j=1}^J$ be a set of part capsules whose states we
want to infer\footnote{We do not need to infer the existence
probabilities of these capsules, since the capsules are value coded and every
capsule can make itself useful by modeling some part of the input. While
simpler objects should ideally require fewer capsules to describe, in this work
we developed our models using a fixed number of part capsules for all
objects.}. Let $R_{ij}\in[0,1]$ be the probability of $\vx^i$
belonging to part $j$. To apply Multi-View Agreement, we need
a feature-voting network~$F$, pose-voting network~$Q$, and
decoding network~$G$.

We architect $Q$ and $F$ to be similar to a PointNet~\citep{pointnet},
where a function embeds the points, weights them by their membership $R_{ij}$ to the part, before
max-pooling them:
\vspace{-0.25em}
\begin{align}
Q(X|{\vz^j}, R) &= Q_{\text{project}}(\text{maxpool}_i R_{ij} Q_{\text{embed}}(\vx^{i}|{\vz})), \label{eq:Q} \\
F(X|{\vz^j}, R) &= F_{\text{project}}(\text{maxpool}_i R_{ij} F_{\text{embed}}(\vx^{i}|{\vz})). \label{eq:F}
\end{align}
Both the embedding
and projection networks consist of 3 ResNet blocks. 
The output for $Q$ is a 7-D vector, which is interpreted as a pose transformation
by normalizing the last 4 dimensions to lie on a unit sphere and multiplying
them by $-1$ if the first element is negative. $F$ outputs the percept.

\begin{figure}[t!]
\begin{subfigure}{.577\linewidth}
    \includegraphics[width=\linewidth]{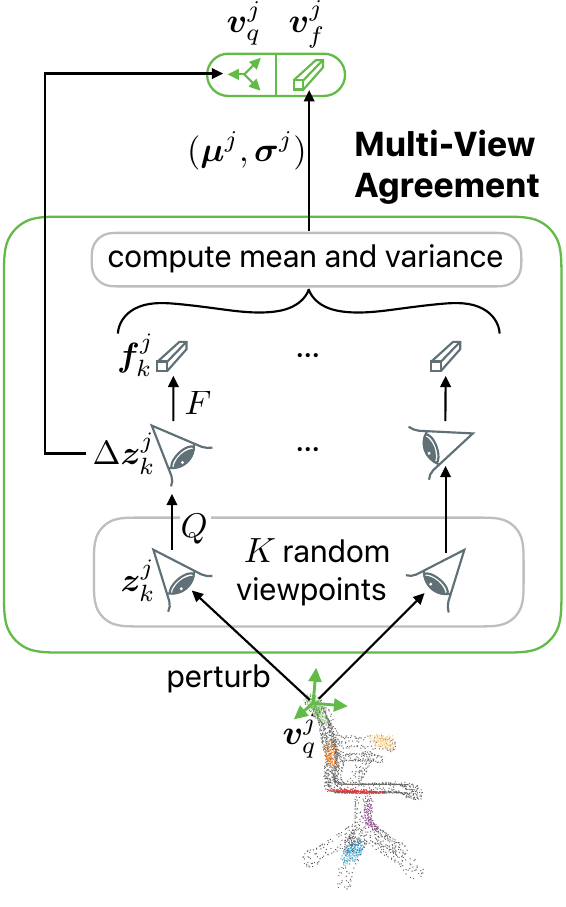} 
\caption{\small PointsToParts Encoder.}
\label{fig:encoder}
\end{subfigure}%
\begin{subfigure}{.4035\linewidth}
    \includegraphics[width=\linewidth]{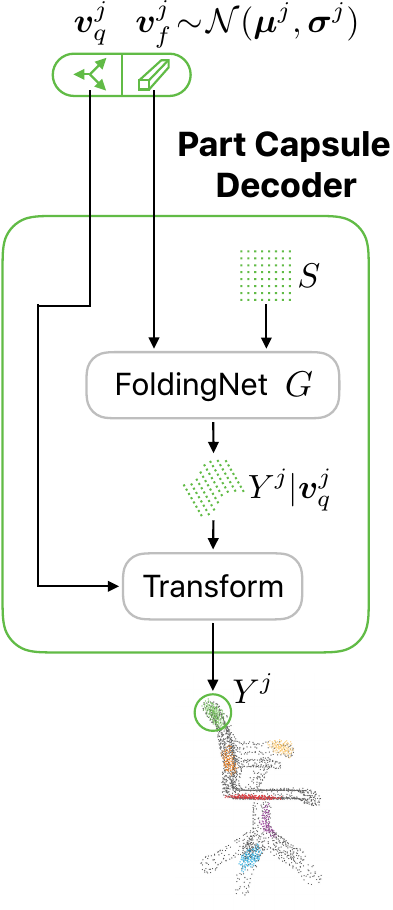} 
\caption{\small PartsToPoints Decoder.\!}
\label{fig:decoder}
\end{subfigure}
\label{fig:pointstoparts}
  \caption{\small PointsToParts Autoencoder. {\bf Left}: Points are encoded into part capsules using Multi-View Agreement. {\bf Right}: Part capsules are decoded into points.}
\end{figure}

The PointsToParts encoding process is described in the {PointsToPartsEncoder}
Procedure of \Algref{alg:pointstoparts} and illustrated in
\Figref{fig:encoder}. The procedure starts with some initial part capsules and
outputs updated part capsules. Part capsules are initialized by setting their
translation components using farthest-point sampling on $X$, their rotation
components randomly, and their feature components to $\vzero$. A random
perturbation (translation and rotation) is applied to the current estimate of
$\vv_q^j$ to generate viewpoints.

\begin{algorithm}[t]
  \caption{PointsToParts}
  \label{alg:pointstoparts}
  \begin{algorithmic}[1]
    \Require 3D Point Cloud $X$, Initial Part Capsules $V$,\ \ \ \ \ \ \ \ \ \ \ \mbox{\ } \mbox{\ \ \ \ \ \ \ \ \ \ Number of iterations $T$}
    \For{$t\leftarrow 1$ to $T$}
      \State $\{Y^j\}_{j=1}^J \leftarrow \text{PartsToPointsDecoder}(V)$
      \For{each $i$, $j$}
        \State $b_{ij} \leftarrow \log P(\vx^i|Y^j)$
       \EndFor
      \State $R_{ij} \leftarrow \softmax_j b_{ij}$
      \State $V \leftarrow \text{PointsToPartsEncoder}(X, R, V)$
      \EndFor
    \State \textbf{return} $V$\vspace{1em}  
    \Procedure{PointsToPartsEncoder}{$X, R, V$}
    \For{$j\leftarrow 1$ to $J$}
      \For{$k\leftarrow 1$ to $K$}
          \State $\vz_k^j \leftarrow \vv_q^j \cprod \vr, \ \ \ \vr \sim$ random perturbation
          \State $\Delta \vz_k^j \leftarrow Q(X|{\vz_k^j}, R)$
	          \State $\vf^j_k \leftarrow F(X|{\vz_k^j \cprod \Delta \vz_k^j}, R)$
        \EndFor
      \State $\vmu^j, \vsigma^j \leftarrow\text{compute mean and variance of } \{\vf_k^j\}$
      \State $\vv_q^j \leftarrow \vz_1^j \cprod \Delta \vz_1^j$
      \State $\vv_f^j \leftarrow \vmu^j + \vsigma^j * \epsilon, \ \ \epsilon \sim \mathcal{N}(0, 1)$
    \EndFor
    \State $V \leftarrow \{(\vv_q^j, \vv_f^j)\}_{j=1}^J$
    \State \textbf{return} $V$
    \EndProcedure\vspace{1em}
    \Procedure{PartsToPointsDecoder}{$V$}
      \For{$j\leftarrow 1$ to $J$}
        \State $S \leftarrow$ $M$ samples from Unif($[-0.5, 0.5]^2$)
        \State $Y^j = \{\vv_q^j \pointrot \vy | \vy = G(\vv_f^j, s), s \in S\}$
      \EndFor
      \State \textbf{return} $\{Y^j\}_{j=1}^J$
    \EndProcedure
    \end{algorithmic}
\end{algorithm}

The FoldingNet~\citep{foldingnet} is used as the decoding network $G$, where
a surface is represented using a feature vector $\vv_f \in \mathbb{R}^D$, that
describes how a 2D unit square can be folded into that surface. The surface
represented by any feature $\vv^j_f$ can be decoded using a neural network,
$G:(\mathbb{R}^D \times \mathbb{R}^2) \rightarrow \mathbb{R}^3$, that
maps $\vv^j_f$ concatenated with 2D points sampled from a unit square $S$ to 3D
points. The pose $\vv^j_q$ is used to transform the generated 3D surface
to the global frame, as shown in \Figref{fig:decoder}. The {PointsToPartsDecoder} Procedure in \Algref{alg:pointstoparts} describes this in detail.
The union of the surfaces generated from each part capsule $Y = \cup_{j=1}^J Y^j$ is the reconstruction output.
To train the model, the (squared) Chamfer distance between $Y$ and the input point cloud $X$ can now be computed:\\[-1.25em]
\begin{align*}
  \Ls=d_{\text{Chamfer}}(X, Y)=&\frac{1}{|Y|}\sum_{\vy \in Y} \text{min}_{\vx \in X} ||\vx - \vy||^2\\[-0.25em]
  &\!+\frac{1}{|X|}\sum_{\vx \in X} \text{min}_{\vy \in Y} ||\vx - \vy||^2,\\[-2em]
\end{align*}
where loss is minimized using ADAM \citep{adam} to train the neural networks $F$, $Q$, and $G$.

The routing probabilities $R_{ij}$ are estimated through an iterative process.
$\vx^i$ should belong to $\vv^j$, if the point is well explained by the generated part surface $Y^j$ (i.e. it has high log-likelihood under a Gaussian distribution whose mean is defined by $Y^j$). This can be approximated by finding the point in $Y^j$ that is most likely to have generated $\vx^i$:
\begin{align*}
  b_{ij} \equiv \log P(\vx^i | Y^j)\propto-\min_{\vy \in Y^j}\left(\frac{||\vx^i - \vy||^2}{\sigma_j^2} + \log (\sigma_j)\right),
\end{align*}
where $\sigma_j$ is a standard deviation that describes how uncertain the model is about the surface. We tried learning $\sigma_j$ as the output of the
decoding network $G$, but we found that the model works just as well if we simply set it
to a constant ($\approx\!e^{-6}$ in our experiments).
These log probs are used to compute the routing probabilities $R_{ij} =\text{softmax}(b_{i})$. 

Altogether, the PointsToParts Autoencoder works iteratively, as described in \Algref{alg:pointstoparts}. Given $X$ and an initial $V$, the decoder and
encoder are run iteratively to route the points to parts.  We found that 3
iterations of these steps are sufficient to reach convergence for $R$.

\subsection{PartsToObject Autoencoder}
\label{sec:oneobject}

\begin{figure*}[t!]
\begin{center}
\includegraphics[width=\linewidth]{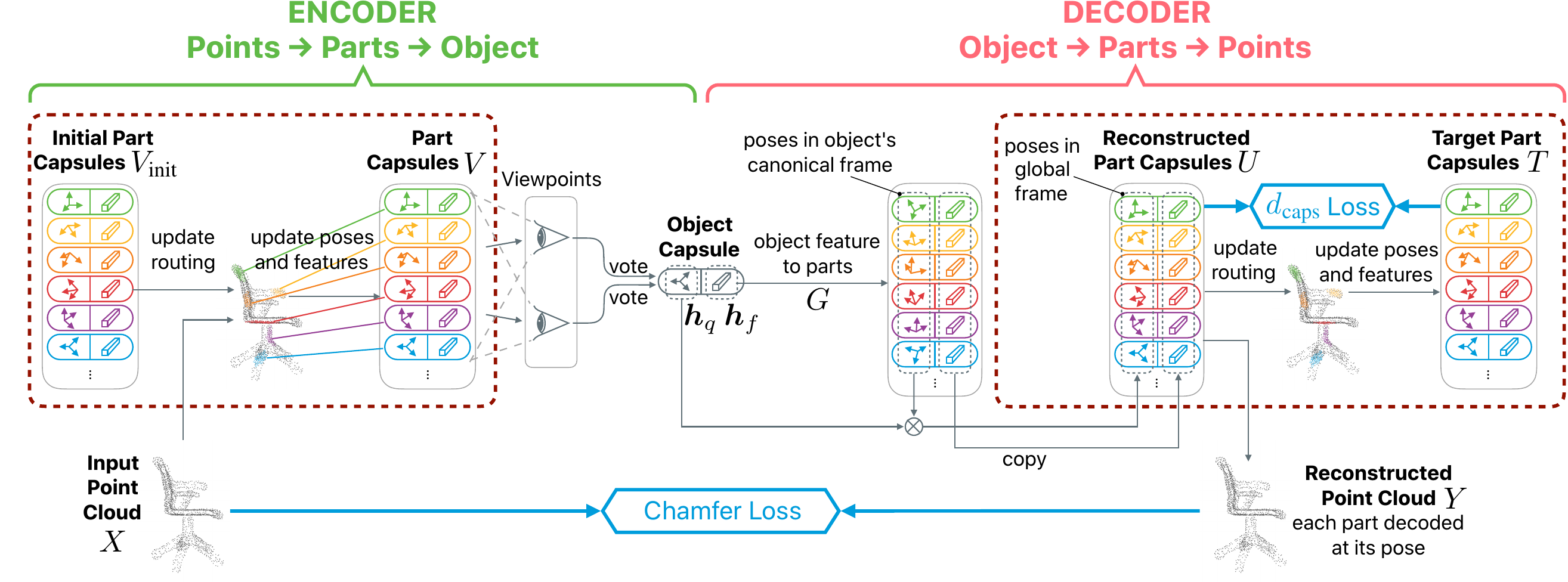}  
\end{center}
    \vspace{-1em}
  
  \caption{\small Object Capsule Autoencoder. Point Cloud $X$ and initial part
  capsules $V_{\text{init}}$ are used to compute part capsules $V$ using the
  PointsToParts model (denoted by the red dashed box). Multi-view agreement is used to compute the object
  capsule $\vh=(\vh_q, \vh_f)$. $\vh_f$ is decoded into part capsules (in their canonical pose). These are transformed into the global frame by applying $\vh_q$, to produce $U$. This is used to output the reconstructed point cloud $Y$ using the decoder from
  the PointsToParts model. In addition, refined part capsules $T$ are computed using the PointsToParts model starting with $U$ as initial part capsules.
  $T$ is used as the regression target for $U$.
  }

  \label{fig:oneobjectautoencoder}
\end{figure*}

A second model is used to construct a single object-level capsule from part-level capsules.
This time, Multi-View Agreement acts on part capsules instead of points.
The process is illustrated in \Figref{fig:oneobjectautoencoder} and in \Algref{alg:pointstoobject}. Starting
from the input point cloud $X$ and initial set of part capsules $V_{\text{init}}$,
an updated set of part capsules $V=\{(\vv_q^j, \vv_f^j)\}_{j=1}^J$ is obtained
using the PointsToParts Autoencoder (\Algref{alg:pointstoparts}).
Now, Multi-View Agreement is applied on $V$. $\vh_q$ is initialized to a random pose and $\vz_k$ are set by randomly perturbing it. The form of the pose voting
network $Q$ and feature voting network $F$ are again similar to PointNet:\\[-1.25em]
\begin{align*}
  Q(V|{\vz}) &= Q_{\text{project}}(\text{maxpool}_j Q_{\text{embed}}(\vv^j|{\vz})),\\[-0.25em]
  F(V|{\vz}) &= F_{\text{project}}(\text{maxpool}_j F_{\text{embed}}(\vv^j|{\vz})).
\end{align*}

\setlength{\textfloatsep}{1.25em}
\begin{algorithm}[tbh]
  \caption{Object Capsule Autoencoder}
  \label{alg:pointstoobject}
  \begin{algorithmic}[1]
    \Require Input Point Cloud $X$, Initial Part Capsules $V_{\text{init}}$
    \State $V \leftarrow \text{PointsToParts}(X, V_{\text{init}}, 3)$
    \State $\vh_q \leftarrow $ random pose
     \For{$k\leftarrow 1$ to $K$}
        \State $\vz_k \leftarrow \vh_q \cprod \vr, \ \ \ \vr \sim \text{random perturbation}$
        \State $\Delta \vz_k \leftarrow Q(V|{\vz_k})$
        \State $\vf_k \leftarrow F(V|{\vz_k \cprod \Delta \vz_k})$
      \EndFor
      \State $\vmu, \vsigma \leftarrow \text{compute mean and variance of } \{\vf_k\}$
      \State $\vh_f \leftarrow \vmu + \vsigma * \epsilon, \ \ \epsilon \sim \mathcal{N}(0, 1)$
      \State $\vh_q \leftarrow \vz_1 \cprod \Delta \vz_1$
      \State $\left(\vu_q^j|{\vh_q}, \vu_f^j\right)_{j=1}^J \leftarrow G(\vh_f)$
      \State $U \leftarrow \left(\vu_q^j, \vu_f^j\right)_{j=1}^J$
      \State $T \leftarrow \text{PointsToParts}(X, U, 3)$
      \State $Y \leftarrow \text{PointsToPartsDecoder}(U)$
      \State Compute Loss $\Ls$ \Comment{From \Eqref{eq:objectloss}}
  \end{algorithmic}
\end{algorithm}

The input to the embedding networks is the concatenation of the relative pose
and feature components of the part capsule $\vv^j|{\vz} = (\vz^{-1} \cprod
\vv^j_q, \vv^j_f)$. Routing probabilities are not required when modeling a single object, since all parts would route to that
object. When dealing with multiple objects, routing probabilities can be incorporated similarly to \Eqref{eq:Q} and \Eqref{eq:F}.
The decoder $G$ differs from the FoldingNet Decoder, where we previously capitalized on the fact that a
deformed surface and a unit square share the same topology. However, it is
not obvious how to define a shared geometric topology for generic objects.
A sphere might be a good candidate, but it would not be able to model
objects that have holes. Therefore, we designed $G$ to
directly output a predetermined number of part capsules\vspace{-0.75em}
\begin{align}
  \left(\vu_q^j|{\vh_q}, \vu_f^j\right)_{j=1}^J = G(\vh_f) \label{eq:objectdecoder},
\end{align}
where $\vu_q^j|{\vh_q}$ is the part's pose in the object's canonical frame,
and $\vu_f^j$ is the part's feature. The part's pose is then composed with
the object's pose, $\vh_q$, to obtain the part's pose in the global frame.
These poses, along with the features, $\vu_f^j$, define the set of
reconstructed part capsules $U = (\vu_q^j, \vu_f^j)_{j=1}^J$ decoded from
object capsule $\vh$.

Computing a reconstruction loss between $U$ and $V$ is not
straightforward as the decomposition of an object into parts is
not unique -- there are infinitely many ways to divide an object into parts and assign poses --
and the PointsToParts Autoencoder may have come up with any one of them (depending on
$V_{\text{init}}$). However, we want $\vh_f$ to be the same no matter how the part
decomposition was done since $\vh_f$ is meant to be a pose-invariant
representation of the object that should only depend on the object's identity.
Therefore, we want all possible $V$'s for the same object to map to the
same $\vh_f$, which through \Eqref{eq:objectdecoder} produces a unique
(canonical) part decomposition $U$.  This makes the desired mapping from $V$ to $U$
many-to-one, so a straightforward regression of $U$ to $V$ (using some form of
a Chamfer distance) does not work. 
Instead, regression targets $T$, are obtained by running a few iterations of the PointsToParts Autoencoder, starting with $U$. This
finds capsule states that are close to, but better than $U$,
according to the PointsToParts Autoencoder.
Intuitively, this
can be seen as the high-level capsule telling the low-level capsules \emph{how}
to view the object, and the low-level capsules then coming up with a local
refinement of that suggestion, which provides the learning signal for the
high-level capsules.
For regression, a distance function between capsules is defined:\\[-1.25em]
\begin{align*}
d_{\text{caps}}(\vv, \vu)=&||\vv_q[t] - \vu_q[t]||^2 + 1 - \left(\langle \vv_q[r],\vu_q[r] \rangle\right)^2 \\[-0.25em]
&\!+ ||\vv_f - \vu_f||^2,\end{align*}
where $\vv_q[t], \vv_q[r]$ represent the translation and rotation components of
the pose. The similarity between quaternions is computed as the \emph{square}
of their dot products.
This takes care of the fact that quaternions have
anti-podal symmetry ($\vr$ and $-\vr$ represent the same rotation).  Given
reconstructed part capsules $U=\{\vu^j\}_{j=1}^J$ and target part capsules
$T=\{\vt^j\}_{j=1}^J$, the regression loss function is then\\[-1.25em]
\begin{align*}
  \Ls_{\text{part}} = \sum_{j=1}^J d_{\text{caps}}(\vu^j, \vt^j).\\[-2em]
\end{align*}

Another learning signal is provided through the reconstruction of the input
point cloud $X$. To do this, we decode $U$ using the PointsToPartsDecoder from
\Algref{alg:pointstoparts} to obtain reconstructed points $Y$ and use Chamfer
distance as the loss function. The overall loss function for the object
autoencoder model is:\\[-1.25em]
\begin{align}
  \Ls = \Ls_{\text{part}} + \reg d_{\text{Chamfer}}(X, Y),\label{eq:objectloss}
\end{align}
where $\reg$ is a hyperparameter tuned on a validation set.

The model is learned by first training the PointsToParts Autoencoder, and then training the PartsToObject Autoencoder, while keeping the first model fixed.

\section{Experiments}
We design experiments to validate two key properties that the learned
object representation should have:
(1) the pose component should be pose-equivariant, and
(2) the feature component should represent object identity (i.e. be pose-invariant).
The first is evaluated using pose alignment from pairs of arbitrarily rotated objects, and the
second, using object retrieval. We also include an ablation study and provide a number
of qualitative visualizations that show the learned representation.

In all our experiments, we train our model on the ShapeNet Core55
dataset~\citep{ShapeNet}, a 55-object-category dataset with 57,448 CAD models,
each uniformly sampled to 2048 3D points.  For evaluation, we used the
ModelNet40~\citep{ModelNet40} dataset, which consists of CAD models from 40
object classes. In particular, we used the same subset of 2,468 test objects
as~\citet{pointnet}. See \Appref{app:dataset} for more details.  During
training, each ShapeNet object is modeled with 16 part capsules (each having an
8-D feature), while the entire object is modeled with a single object capsule
(1024-D feature).  The embedding, projection, and decoder networks consist of
ResNet blocks (details in \Appref{app:pointstoparts} and \Appref{app:partstoobject}).  This unsupervised model is then used {\em without additional
fine-tuning} to extract representations for ModelNet40 objects.  We operate in
this transfer learning setting to simulate a situation where the model is
applied on generic objects that are not known at training time.


\subsection{Pose Equivariance}
We analyze the equivariance of the object capsule's
pose at two levels: (1) by directly changing the part capsules' pose, and (2) by
changing the points' pose. The first setting is designed to specifically test
the PartsToObject model. This setting simulates a situation where changes in
the pose of the object are perfectly captured by the PointsToParts model in
terms of changes in the part capsule's pose only (keeping the part features
invariant) and allows us to evaluate the second layer independent of the
first.

\paragraph{Equivariance w.r.t parts' pose.}
If our model can recover the canonical pose of an object, it should be
possible to align any two views by computing the relative
transformation between the recovered poses.
To test if the proposed Multi-View Agreement method can do this, we do the following experiment.
First, we encode a given test object as part capsules $V_1$.
Next, we apply an arbitrary rotation $\vr$ to the pose components of $V_1$ to generate part capsules $V_2$, where the rotation axis is sampled randomly in $\mathbb{R}^3$ and rotation angle is sampled uniformly in $[-\pi, \pi]$.
We recover the object poses $\vh_1$ and $\vh_2$ for each set of part capsules independently.
If the model is working perfectly, the relative transform $\hat{\vr} = \vh_2 \cprod \vh_1^{-1}$ should be equal to $\vr$.
Subsequently, we compute the distance between the estimated and groundtruth rotation quaternions $d(\vr, \hat{\vr}) = 2\cos^{-1}\left(|\langle \vr, \hat{\vr}\rangle|\right)$, which is the angle encoded by the quaternion $\vr^{-1}\hat{\vr}$. Rotation error is divided by $\pi$ to bound it in $[0, 1]$.
Finally, we evaluate the average rotation error on likely-asymmetric object classes (chair, bed, sofa, toilet, monitor), following~\citep{qenet}, from the ModelNet10 dataset (a subset of the ModelNet40).

\begin{figure}[t]
\begin{subfigure}{\linewidth}
\begin{center}
    \includegraphics[width=\linewidth, trim=112 166 206 165, clip]{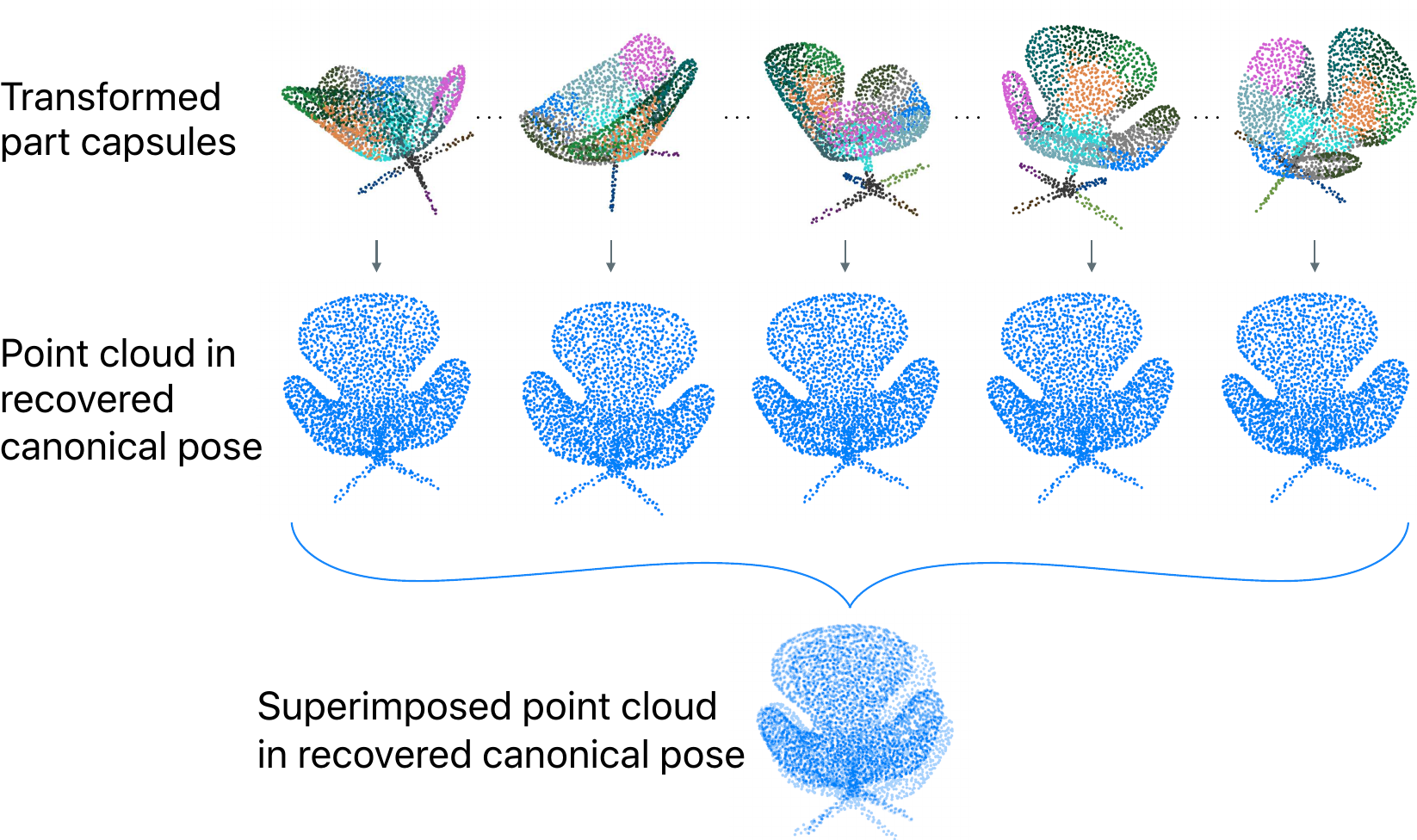} 
\caption{\small Recovered canonical pose from parts that have been transformed in different ways. The superimposed point cloud visually indicates the stability of the recovered canonical pose.\!\!\!}
\end{center}
\label{fig:canonical_pose1}
\end{subfigure}
\vspace{1em}
\begin{subfigure}{\linewidth}
    \includegraphics[width=\linewidth, trim=130 84 186 80, clip]{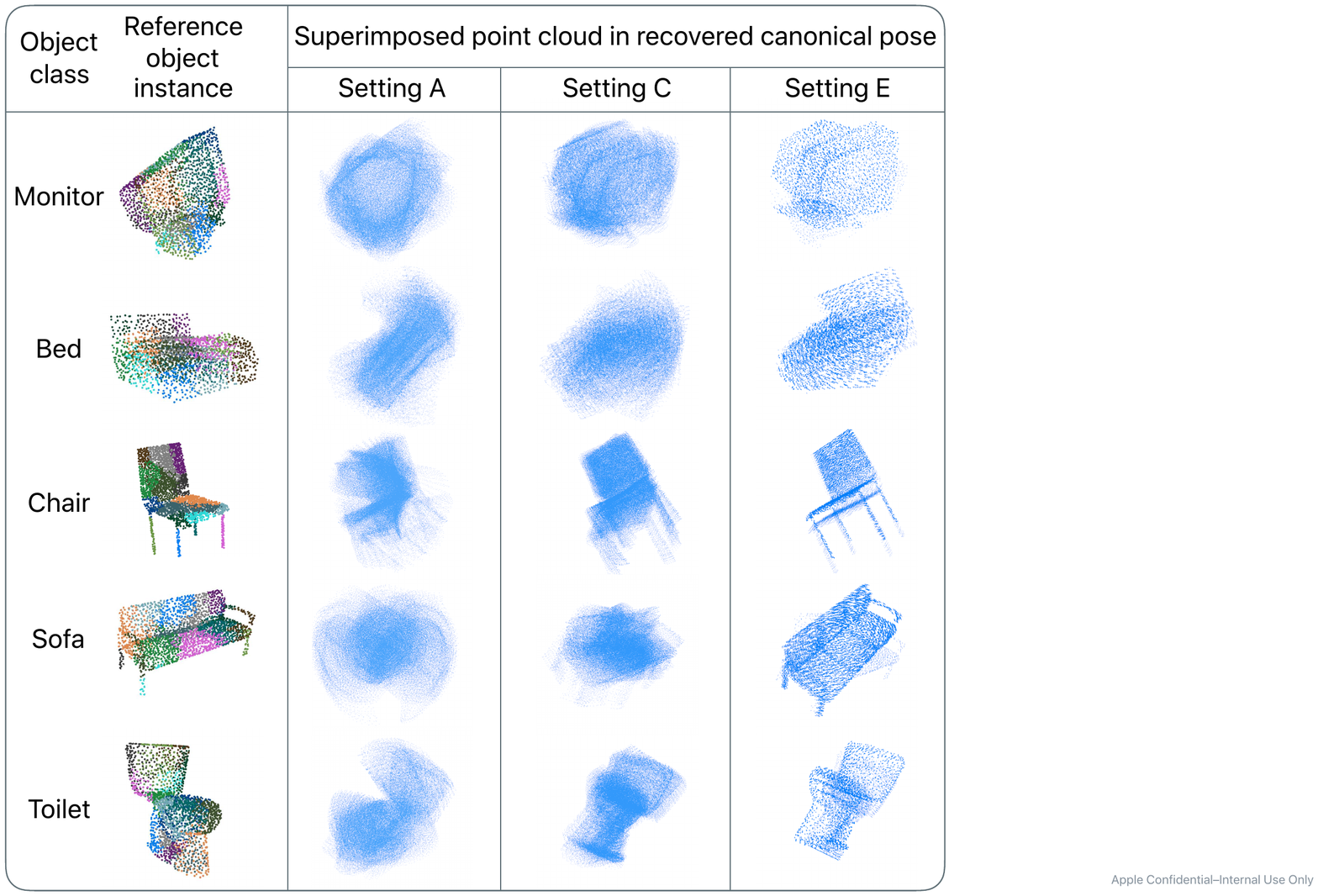} 
\vspace{-1em}
\caption{\small Examples of superimposed point clouds in recovered canonical poses across three different experimental settings (see \Tabref{tab:align}). Setting F consistently produces the most visually stable recovered canonical poses across the set of 50 transformations.}
\label{fig:canonical_pose2}
\end{subfigure}
\vspace{-0.5em}
  \caption{\small Visualization of pose equivariance results. Each object is encoded into part capsules. The part capsules are then transformed by rotating about a random axis. A canonical pose is inferred from the rotated capsules, and the object is visualized in this canonical pose. If the visualization is stable across different rotations, then the object representation is pose equivariant.}
\label{fig:canonicalvieweval}
\end{figure}

We do an ablative study to quantify the impact of the design choices made in our
model.  \Tabref{tab:partposeequi} shows the results of our experiments, 
where we vary different aspects of our model, such as the number of
viewpoints and amount of perturbation added \emph{only during training}. 
At test time, only one viewpoint is used and no perturbation is added, to
ensure a fair comparison. Our first model (Setting A) has only
one view that is allowed to vote to transform itself, record a percept and
reconstruct from that. Viewpoint perturbation is irrelevant here, since
there is only a single randomly-initialized viewpoint. This model has a
high pose error rate of 0.259, even though the autoencoder's reconstruction error was similar to the other models.
This shows that a model that can reconstruct well 
might not have necessarily learned a canonical pose. Having two views (Setting B) that have to agree with each
other during training, drops the error rate to 0.135. Viewpoint perturbation of [-45$^{\circ}$, 45$^\circ$] about a random axis was used. Increasing the number of views to 4 (Setting C)
yielded a similar result. 
If we gradually ramp up the noise injected in the view perturbations from 45$^{\circ}$ to
180$^{\circ}$ as training progressed, the error reduces further (Setting D). Finally, if instead of doing just one
step of pose voting, we perform 3 steps (each time starting from the previous
viewpoint), we see a significant improvement to 0.023 (Setting E). This shows that the pose voting
network can move random viewpoints to canonical ones and is able to do so with
increasing accuracy when applied repeatedly.

For each object, we pick a random axis and rotate the object about it by an angle going from -180$^{\circ}$ to
180$^{\circ}$ in 50 steps. The point cloud viewed in the canonical pose
discovered by model is superimposed and visualized (\Figref{fig:canonicalvieweval}). We observe that the model trained with
multiple views and multiple pose voting steps recovers the most stable canonical pose,
showing that the PartsToObject layer works well due to the components proposed in
our model.  It should be noted that this performance is under the assumption
that the PointsToParts layer was working ideally.

\setlength{\tabcolsep}{12pt}
\begin{table*}[t]
\caption{\small Results on equivariance and invariance w.r.t parts' pose.}
\label{tab:align}
\begin{center}
\begin{tabular}{lcccc}
\hline\\[-1em]
\multirow{ 2}{*}{\bf Setting}  				& \bf Average 		&  \multicolumn{2}{c}{\bf Instance Retrieval} 	&\bf 1-NN\\[-0.1em]
    									&  \bf Rotation Error 	&  \bf Top-1 	&   \bf Top-10  	&  \bf  Classification		\\
\hline \\[-1em]
A. views=1                				&                 0.259 		& 0.286			& 0.386			& 0.485			 			\\ 
B. views=2, noise=45                				&       0.135 		& 0.598			& 0.765			& 0.743			 			\\
C. views=4, noise=45                				&       0.134 		& 0.620			& 0.787			& 0.754			 			\\
D. views=4, noise=[45:180]         			&           0.106 		& 	0.701		& 0.857			& 0.803			 			\\
E. views=4, noise=[45:180], 3 voting steps 	& 	    0.023 		& 	0.943		& 0.959			& 0.960			  			\\
\hline
\end{tabular}
\end{center}
  \label{tab:partposeequi}
\end{table*}

\paragraph{Equivariance w.r.t points' pose.} 
To jointly evaluate pose equivariance of the two model layers, we use the very hard task of aligning two randomly-rotated point clouds of an object without processing both together (i.e. rotating each one to its canonical pose independently).
We compare with two recently proposed deep networks that to do point cloud alignment.
PointNetLK~\citep{pointnetlk} uses a PointNet as an imaging function and applies the Lucas-Kanade approach to align two views,
while QE-Net~\cite{qenet} learns quaternion-valued pose representations by aggregating pose-votes from local regions.
In the experimental setting of~\cite{qenet}, which we also follow, aligning using PCA gets a rotation error of 0.42, PointNetLK get 0.38 and QE-Net gets 0.17.
In the same setting, our model gets 0.16, if we allow 10 trials and pick the transformation that leads to the least Chamfer
distance after alignment. With one trial, performance is similar to PCA (0.42). We found that multiple trials were necessary since
the model suffers from a local-minima problem. Instead of finding a single unique
canonical pose, the model learns a number of attractor regions.
Depending on the initialization of the object pose, it finds one of these canonical poses.
These poses often correspond to principle axes of the input point cloud. Since
the pose-voting and percept-voting networks are jointly learned, it is
likely that the pose-voting network commits to certain canonical poses based on
an imperfect (partially trained) percept function, which only captures the
principle axes to begin with. Even though the percept function becomes more
discerning later in training, the pose votes are not likely to change
drastically, which is often required to go from one direction of the principle
axes to another.  Investigating better training strategies for this model is an
important direction for future work.

\subsection{Pose Invariance Analysis}\label{sec:classification}
The latent variable $\vh_f$ represents the identity of the input
object and should be invariant to its pose. We evaluate whether $\vh_f$
extracted from an object can be used as a retrieval query to find it within a database (test set of ModelNet40) of randomly rotated objects. 
Query matching is done using the L2 distance between the features.
Similar to the pose equivariance experiments, evaluation is done separately for the second
layer 
and the combined model.

\paragraph{Invariance w.r.t parts' pose.} We use the model's variants (Settings
A-E) to extract 1024-D object features from randomly rotated objects in the
database. The part capsules extracted in this process are randomly rotated to
generate the queries to be testing against.  Retrieval accuracies are
computed by checking if the correct object instance is the best match or 
within the top 10 matches. We also check if the best matching object
belongs to the same class as the query (Nearest Neighbor Classification).
From \Tabref{tab:partposeequi}, we see that retrieval
performance has a similar trend as rotation error.
Models trained with multiple views and more voting steps perform better.
The best model has a top-1 accuracy of 94.3\%, top-10 accuracy of 95.9\%, and
1-NN classification accuracy of 96.0\%, indicating that the model almost always
finds the exact instance even among rotated objects.

\paragraph{Invariance w.r.t points' pose.} 
In this setting, we rotate the points, re-encode the points using the part capsule model,
and extract an object feature from the new part capsules.
Here, performance is significantly worse,
with the best model getting top-1 and top-10 accuracies of 7.8\% and 17.9\% respectively.
The model is generally unable to find a canonical view and ends up
representing different views of an object with different feature vectors.
However, the 1-NN classification accuracy is 44\%, which means that although
the model rarely gets the exact instance matched, it frequently matches with other instances in the same class.

\subsection{Visualizing Part Features}
\begin{figure}[t]
  \begin{center}
  \includegraphics[width=\linewidth]{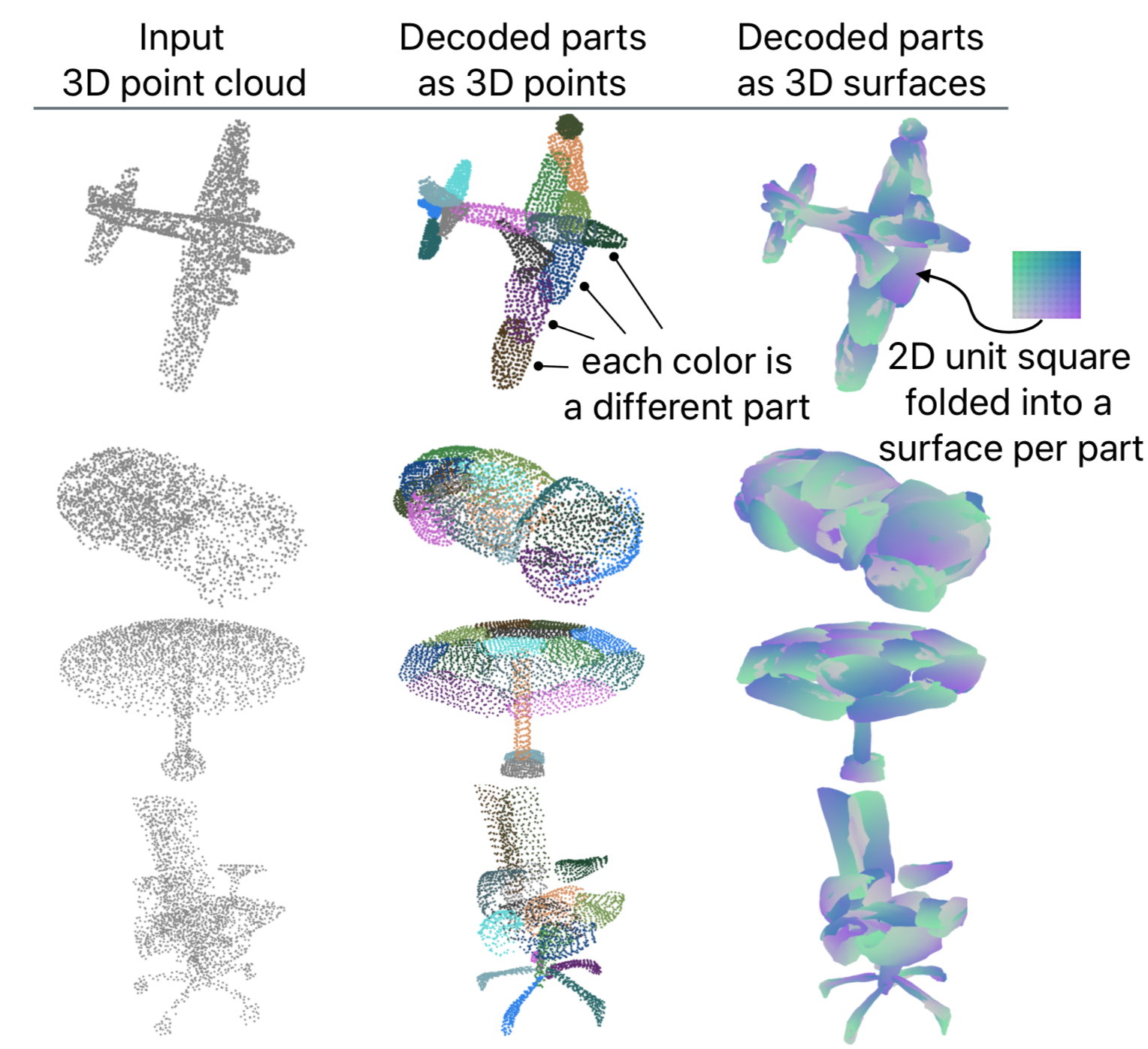}  
  \end{center}
  \caption{\small Visualization of part decompositions obtained by the PointsToParts model.}
  \label{fig:partexamples}
\end{figure}

Our model learns to represent a point cloud as a collection of surfaces
(\Figref{fig:partexamples}).  This visualization shows part capsules obtained
by the PointsToParts model.  We observe that the parts capture different types
of surfaces including sharply curved ones. Each part typically latches onto a
smooth local region. The parts discovered in our model are more local, compared
to parts learned by the 3D Point Capsule Network (3DCapsNet)~\citep{3dcapsnet}.
This is a direct result of representing the pose separately from the feature
that defines the shape of the part.  In the absence of this, the FoldingNet in
3DCapsNet needs to model both the shape and location of the part, leading to
parts that are not effectively constrained to be local. More part
visualizations, including those obtained from the object capsule autoencoder,
are shown in \Appref{app:moreviz}.

\subsection{Discussion}
Our experiments demonstrate some key strengths and
weaknesses of our approach.  Based on experiments with just the PartsToObject
layer, we showed that the Multi-View Agreement algorithm discovers canonical
poses and pose-invariant representations, and effectively solve alignment
and retrieval tasks.  The visualized part features suggests that the
same algorithm when applied to points, can simultaneously do routing and learn
plausible parts.  Our work also highlights a key issue with learning
interpretable object representations, which is how to handle multiple valid
explanations of an object.  In our case, many instantiations of
part capsules explain the points equally well. The second layer must
reconcile these to learn a unique object-level feature and pose.
Having multiple ways to interpret an object is a fundamental property of the
visual world and models that aim to learn object representations must solve it
in some way. In our capsule network design, we solve this by using top-down
information, i.e. the object-level capsule tells the lower-level capsules in
the decoder how to view the points. This decouples the input part decomposition
from the output one. Having this novel decoder design made it possible to train
our networks. The canonical poses discovered by the overall model are still not
entirely unique. However, this work provides some evidence that it is possible
to learn interpretable geometric representations and Multi-View Agreement can
be a useful tool towards that goal.

\section{Conclusion}
We proposed a model that represents parts and objects
as discrete computational entities, called geometric capsules. We developed a
novel Multi-View Agreement algorithm that infers parent capsule poses.
We quantified the degree to which the invariance and equivariance
properties of our learned representation hold in the PartsToObject layer and
in the model as a whole. Our analyses show the benefits of having multiple
votes agree. In future work, we would like to further improve the stability of
the model and apply it to tasks that benefit from using the persistence of
part-whole geometric relationships over time, such as object tracking under
heavy occlusion in point cloud sequences.

{\small
\bibliographystyle{ieee_fullname}
\bibliography{egbib}
}
\clearpage
\appendix
\appendixpage
\section{Details for the PointsToParts Model}
\label{app:pointstoparts}
In this section, we describe the network architecture and training details for
the first layer of our model which encodes a point cloud as a set of part
capsules.

\subsection{Architecture}

\paragraph{Pose-Voting Network}
The pose-voting network $Q$ consists of an embedding network $Q_{\text{embed}}$ and a projection network $Q_{\text{project}}$. Both networks are made of ResNet blocks. A ResNet block takes the following form:
\begin{align*}
  \text{Res}_D(\vx) = \text{Relu}(\vx + W_2\text{Relu}(W_1\vx))
\end{align*}
where $\vx$ is an $N$-dimensional vector, $W_1$ is an $N \times D$ matrix, and $W_2$ is a $D \times D$ matrix.
$Q_{\text{embed}}$ consists of a linear projection of a 3D point into 64-dimensions, followed by a Relu non-linearity, followed by 3 $\text{Res}_{64}$ blocks:
\[
  Q_{\text{embed}}:  3 \rightarrow 64 \rightarrow \text{Relu} \rightarrow \text{Res}_{64} \rightarrow \text{Res}_{64} \rightarrow \text{Res}_{64},
  \]
$Q_{\text{project}}$ takes the pooled embeddings as input and applies the following operations:
\[
  Q_{\text{project}}:  64 \rightarrow 64 \rightarrow \text{Relu} \rightarrow \text{Res}_{64} \rightarrow \text{Res}_{64} \rightarrow \text{Res}_{64} \rightarrow 7
  \]
The 7-D output is interpreted as a pose vote : 3 (translation vector) + 4 (rotation quaternion).

\paragraph{Feature-Voting Network}
The design of $F_{\text{embed}}$ is the same as that of $Q_{\text{embed}}$.
The design of $F_{\text{project}}$ is also the same as $Q_{\text{project}}$,
except that instead of projecting to 7-dimensions, the result is projected to 8
dimensions and is interpreted as a real-valued percept (no non-linearity is
applied).
\begin{align*}
  F_{\text{embed}}:&  3 \rightarrow 64 \rightarrow \text{Relu} \rightarrow \text{Res}_{64} \rightarrow \text{Res}_{64} \rightarrow \text{Res}_{64}\\
  F_{\text{project}}:&  64 \rightarrow 64 \rightarrow \text{Relu} \rightarrow \text{Res}_{64} \rightarrow \text{Res}_{64} \rightarrow \text{Res}_{64} \rightarrow 8
\end{align*}

\paragraph{Decoding Network}
The decoding network $G$ takes the form 
\[
G:  [64 +2] \rightarrow 64 \rightarrow \text{Relu} \rightarrow \text{Res}_{64} \rightarrow 3,
  \]
where $[64+2]$ represents the concatenation of a 64-D feature with a point $(x,y)$ sampled from a unit square.

\subsection{Training}

The model is trained with ADAM using a mini-batch size of 32, starting with a
learning rate of $0.001$ and decaying it by a factor of $0.1$ after 20K and
100K updates. A very low L2 decay of $10^{-7}$ is used for all parameters. The
random viewpoints used in Multi-View Agreement are generated by adding a
perturbation noise that is randomly sampled in $[-45^{\circ}, 45^{\circ}]$.
The input consists of $2048$ points which are translated (by a random vector in
$[-1, 1]^3$) and rotated (about a random axis and random angle $[-180^{\circ},
180^{\circ}]$) as a form of data augmentation. 16 part capsules are used, each
with an 8-dimensional percept. 4 random views are used per capsule. Each part
capsule is decoded by sampling $M=256$ points in a unit square. During
training, 3 iterations of routing were done. We found that back-propagating
gradients through the routing iterations did not help performance. So for all
the results in the paper, we trained our model by obtaining the point-to-part
routing $R$ (using 2 iterations), doing one application of the encoder, one of
the decoder, and backpropagating through those last two applications only.

\section{Details for the PartsToObject Model}
\label{app:partstoobject}

In this section, we describe the network architecture and training details for
the model that learns to represent an object as a single capsule.

\subsection{Architecture}
\paragraph{Pose-Voting Network} The network $Q_{\text{embed}}$ takes the following form,
\begin{align*}
  Q_{\text{embed}}: [D+7] \rightarrow 1024 \rightarrow \text{Relu} \rightarrow \text{Res}_{1024} \times 3
\end{align*}
where $D=8$ is the dimensionality of each part capsule's feature component,
$[D+7]$ represents the concatenation of the feature and pose components and $\text{ResNet}_D \times 3$ represents 3 $\text{ResNet}_D$ blocks applied in sequence (similar to the first layer). $Q_{\text{project}}$ takes the form:
\begin{align*}
  Q_{\text{project}} : 1024 \rightarrow 1024 \rightarrow \text{Relu} \rightarrow \text{Res}_{1024} \times 3 \rightarrow 7
\end{align*}

\paragraph{Feature-Voting Network}
The design of $F_{\text{embed}}$ is the same as that of $Q_{\text{embed}}$.
The design of $F_{\text{project}}$ is also the same as $Q_{\text{project}}$,
except that instead of projecting to 7-dimensions, the result is projected to 1024
dimensions and is interpreted as a real-valued percept (no non-linearity is
applied).
\begin{align*}
  F_{\text{embed}}:&  [D+7] \rightarrow 1024 \rightarrow \text{Relu} \rightarrow \text{Res}_{1024} \times 3\\
  F_{\text{project}}:&  1024 \rightarrow 1024 \rightarrow \text{Relu} \rightarrow \text{Res}_{1024} \times 3 \rightarrow 1024
\end{align*}

\paragraph{Decoding Network}
The decoding network $G$ maps a 1024-dimensional object percept to $J=16$
parts. We found that it worked best to train a separate decoder for each part
capsule. Therefore, $G$ consists of $\left(G_1, G_2, \ldots, G_J\right)$, where
\begin{align*}
  G_j : 1024 \rightarrow 256 \rightarrow \text{Relu} \rightarrow \text{ResNet}_{256} \times 4 \rightarrow [D+7].
\end{align*}

\subsection{Training}

The model is trained with ADAM using a mini-batch size of 32, starting with a
learning rate of $10^{-4}$ and decaying it by a factor of $0.1$ after 50K and
100K updates. A low L2 decay of $10^{-7}$ is used for all parameters. The
random viewpoints are generated by adding perturbation noise that is randomly
sampled in $[-\theta, \theta]$, where $\theta$ starts off at
$45^{\circ}$ and is increased linearly during training to $180^{\circ}$
starting at 10K updates and ending at 50K. 4 random views were used, and 2 also
gave similar results. The weight $\lambda = 0.01$ worked well for combining the
Chamfer loss and $\mathcal{L}_{\text{part}}$. We found that $\lambda=0$ actually worked
just as good in terms of the pose equivariance/invariance experiments, but
$\lambda=0.01$ led to better reconstructions.  When training the object capsule
layer, the part capsule layer was pre-trained and kept fixed. 3 iterations were
done in the part capsule layer for generating the input capsules as well as for
generating the target capsules.

\section{Datasets}
\label{app:dataset}
ShapeNet~\citep{ShapeNet} was used for training
and ModelNet40~\citep{ModelNet40} for evaluation. For both datasets, we used
pre-processed point clouds provided by Zhao et al.  \citep{3dcapsnet} with their
3D Point Capsule Network work. Each object consists of 2048 points
sampled uniformly at random on the surface of the object. A validation set of
5000 randomly chosen objects was held out from the ShapeNet dataset. The
remaining 52,448 objects were used for training. We used the test set of the
ModelNet40 dataset for evaluating pose invariance (2468 objects). For the pose
equivariance evaluation, we only used the intersection of this test set with
the likely-asymmetric classes from ModelNet10 (bed, chair, sofa, toilet,
monitor) to coincide with the setup from QE-Net \citep{qenet}.

\section{Implementation}
\label{app:implementation}
The model was implemented using PyTorch
\citep{pytorch}. The training was done on a single NVIDIA V100 GPU with 32GB
memory. It took 100K iterations to train the first model (few hours) and 500K
for the second layer (half a day). We are planning to make our implementation
publicly available.

\section{Additional Visualizations}
\label{app:moreviz}

\begin{figure*}
  \centering
  \includegraphics[width=\linewidth, trim=100 420 250 150, clip]{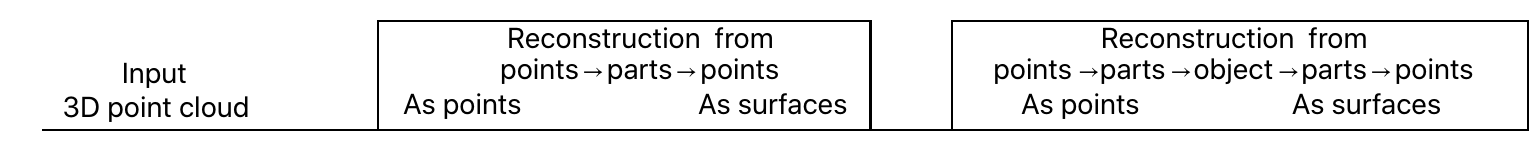}  
  \includegraphics[width=\linewidth, trim=0 0 0 150, clip]{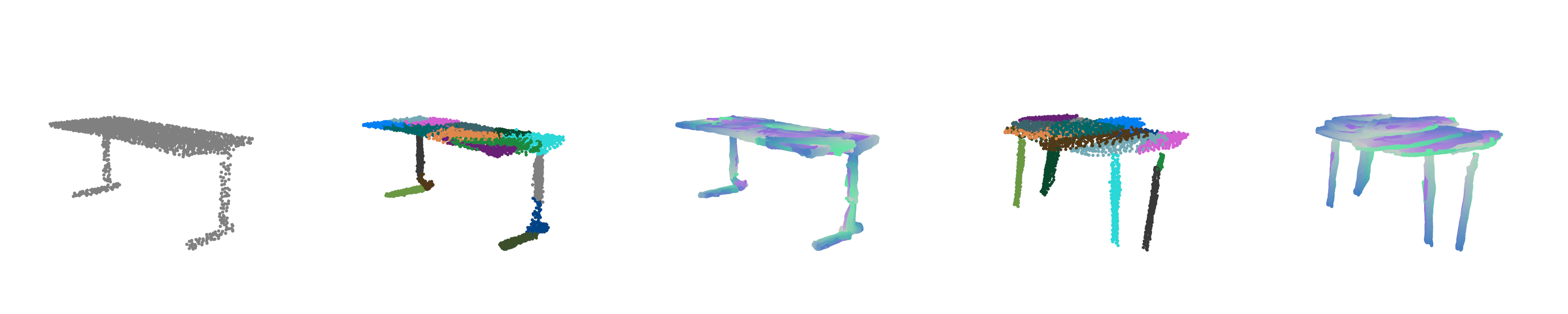}
  \includegraphics[width=\linewidth]{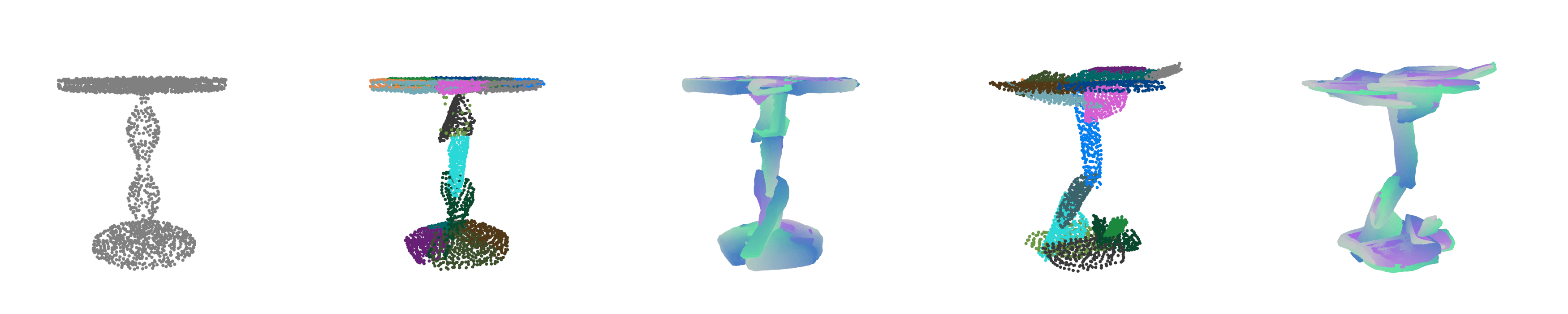}
  \includegraphics[width=\linewidth]{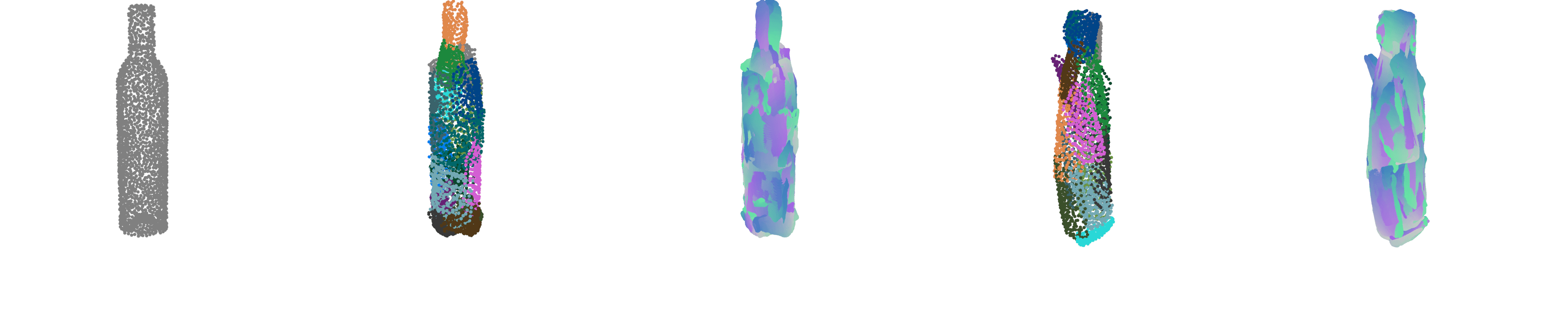}
  \includegraphics[width=\linewidth]{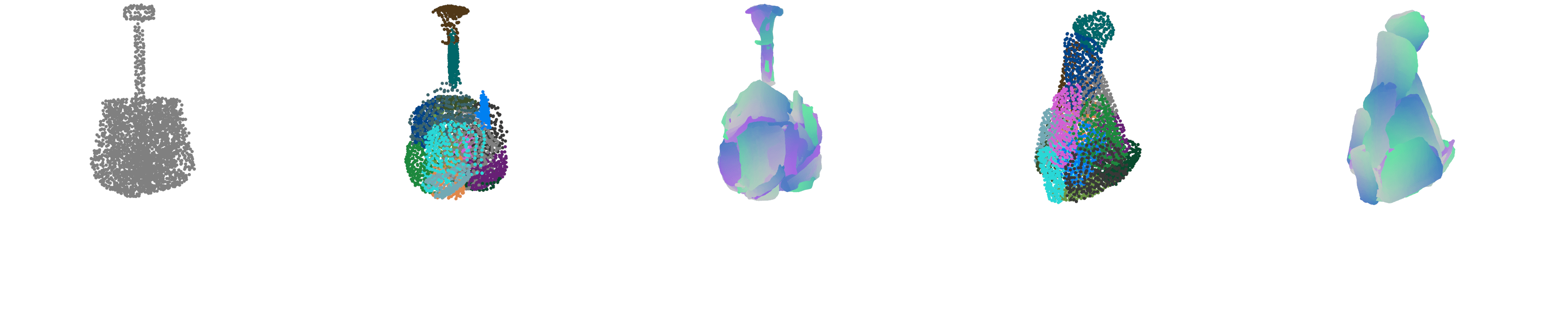}
  \includegraphics[width=\linewidth]{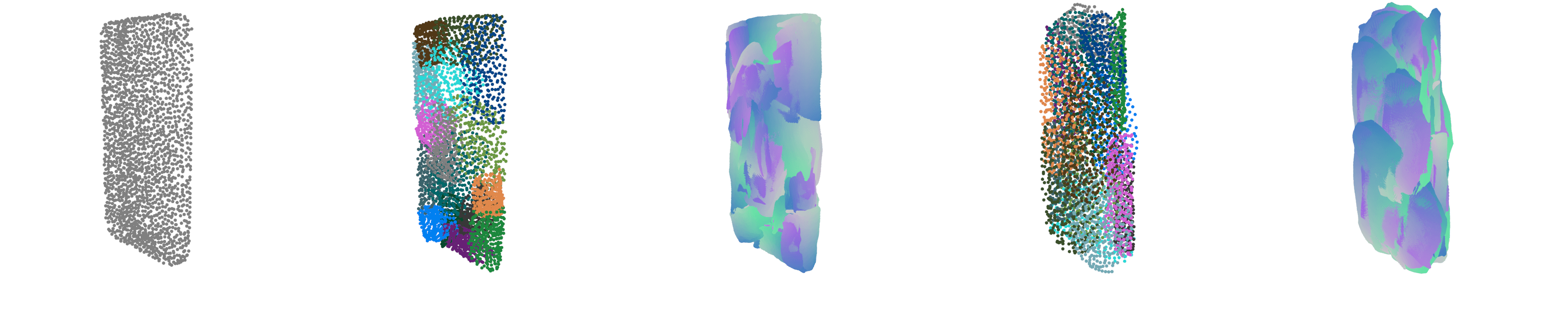}
  \includegraphics[width=\linewidth, trim=0 150 0 0, clip]{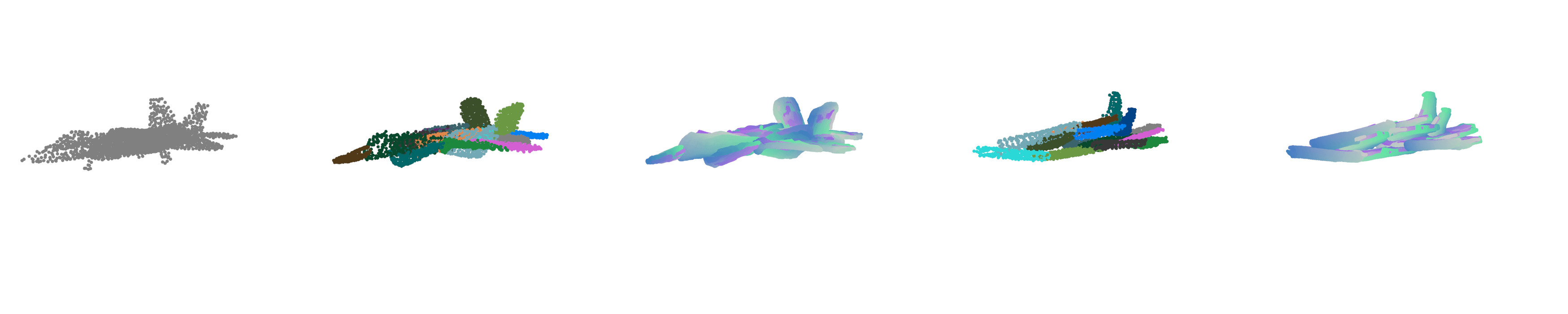}
  \caption{\small Reconstructions from the model. The first column shows the input point cloud. The second column shows reconstructions from the PointsToParts model represented as 3D points. The third column shows the same reconstructions, visualized as 3D surfaces. The fourth and fifth columns show reconstructions from the object-level capsule (as points and surfaces respectively).}
  \label{fig:recon1}
\end{figure*}

\begin{figure*}
  \centering
  \includegraphics[width=\linewidth, trim=100 420 250 150, clip]{images/supp_header.pdf}  
  \includegraphics[width=\linewidth, trim=0 0 0 150, clip]{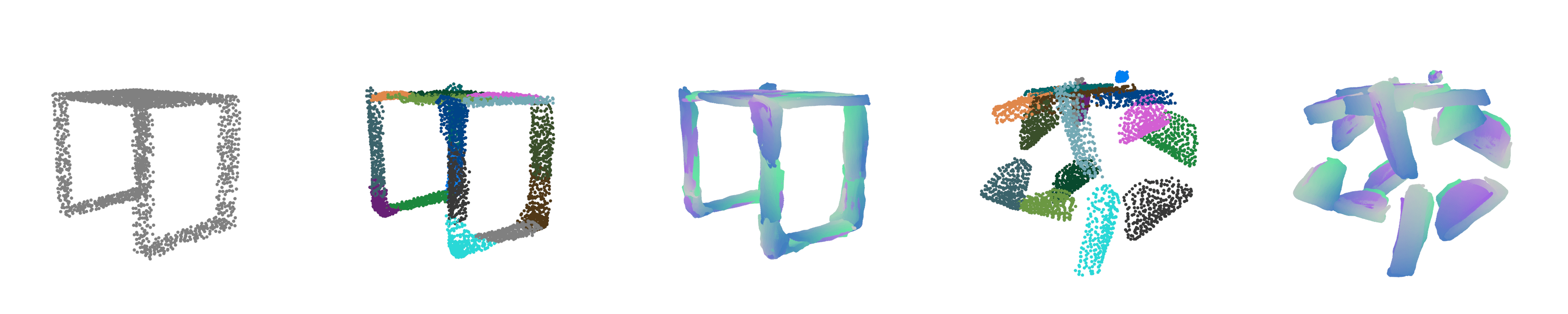}
  \includegraphics[width=\linewidth]{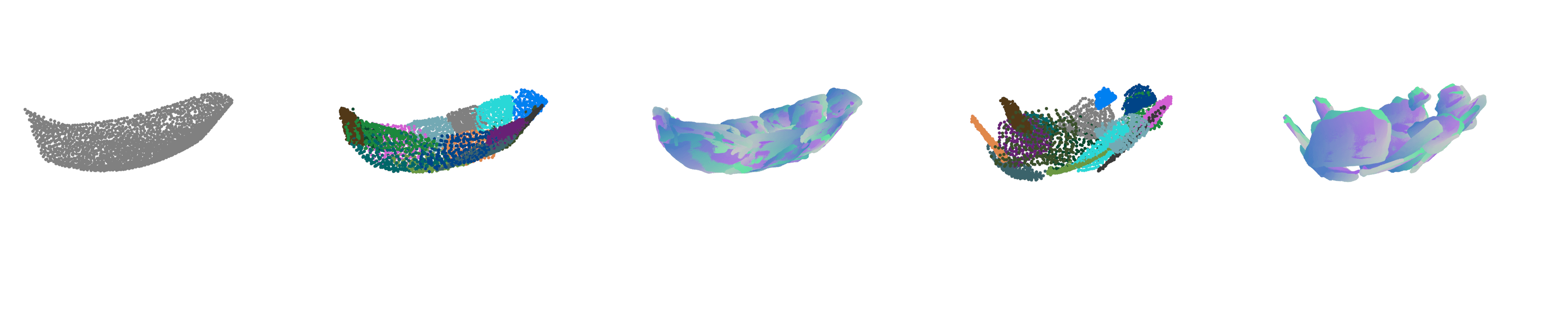}
  \includegraphics[width=\linewidth]{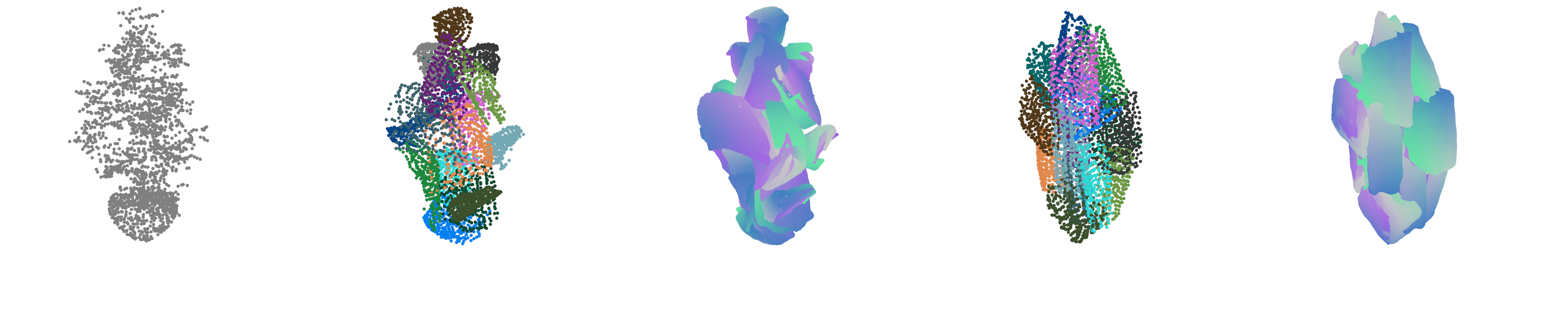}
  \includegraphics[width=\linewidth]{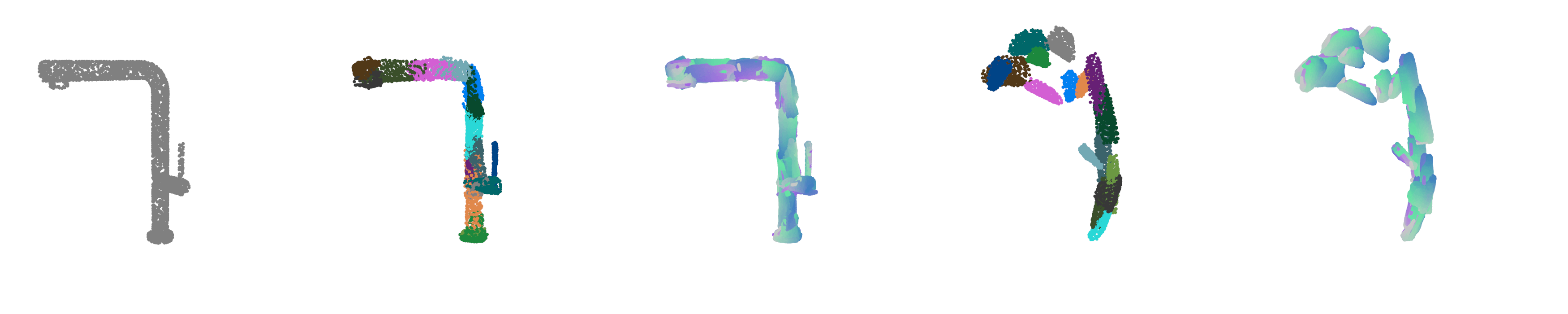}
  \includegraphics[width=\linewidth]{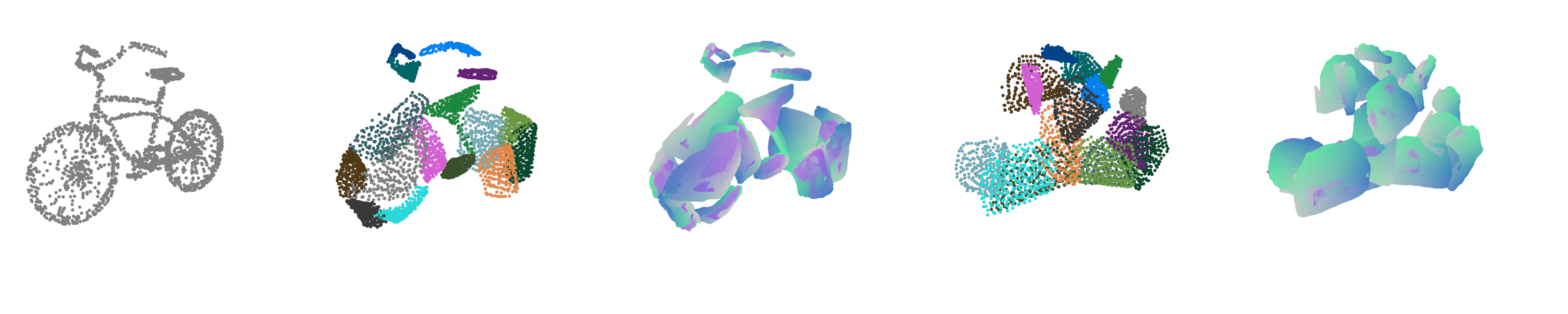}
  \includegraphics[width=\linewidth, trim=0 150 0 0, clip]{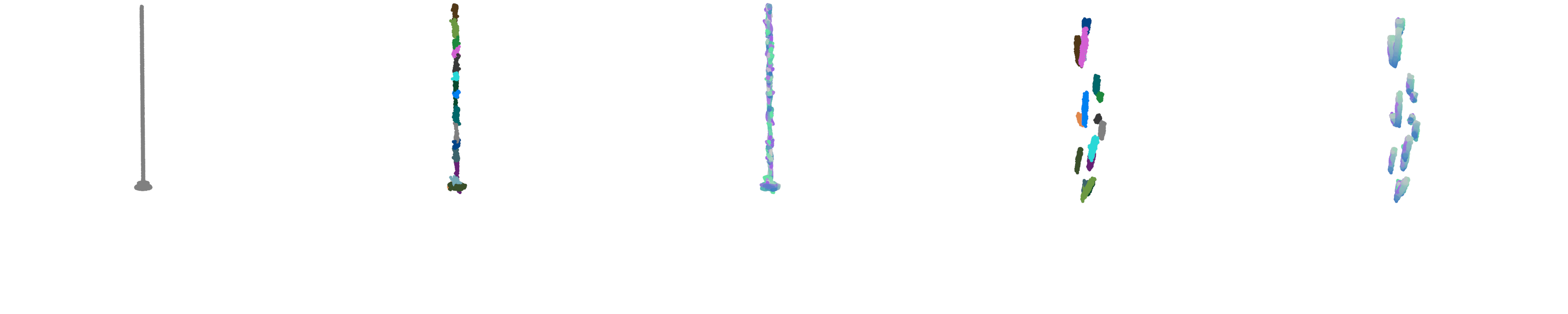}
  \caption{\small Reconstructions from the model. The first column shows the input point cloud. The second column shows reconstructions from the PointsToParts model represented as 3D points. The third column shows the same reconstructions, visualized as 3D surfaces. The fourth and fifth columns show reconstructions from the object-level capsule (as points and surfaces respectively).}
  \label{fig:recon2}
\end{figure*}

\Figref{fig:recon1} and \Figref{fig:recon2} show reconstructions obtained by
our model. We visualize reconstructions from a single layer
autoencoder (points $\rightarrow$ part capsules $\rightarrow$ points), as well
as reconstructions from the full two-layer model (points $\rightarrow$ part
capsules $\rightarrow$ object capsule $\rightarrow$ part capsules $\rightarrow$
points). These objects are taken from the validation set that was used during
training (subset of ShapeNet). The first example in \Figref{fig:recon1} shows
an interesting failure case, where the object capsule model substituted four
legs in the place of the two present in the input. Overall, the reconstructions
from the single-layer model look much better than the two-layer model. This is
because the single layer model is only modeling the small parts individually,
while the two-layer model has to model the appearance of all the parts and their
poses.


\end{document}